\documentclass[sigconf,nonacm]{acmart}

\AtBeginDocument{%
  }

\usepackage{algorithm}
\usepackage{algorithmicx}
\usepackage{algpseudocode}
\usepackage{subcaption}
\usepackage{multirow}
\usepackage{enumitem}
\usepackage[normalem]{ulem}

\algdef{SE}[DOWHILE]{Do}{doWhile}{\algorithmicdo}[1]{\algorithmicwhile\ #1}%

\begin{document}

\title{A Robust Clustered Federated Learning Approach for Non-IID Data with Quantity Skew}

\author{Michael Ben Ali}
\affiliation{
  \institution{Université Toulouse UT3, IRIT, CNRS}
  \city{Toulouse}
  \country{France}
}
\email{michael-eddy.ben-ali@irit.fr}

\author{Imen Megdiche}
\affiliation{
  \institution{INU Champollion ISIS Castres, IRIT, CNRS}
  \city{Castres}
  \country{France}
}
\email{imen.megdiche@irit.fr}

\author{André Peninou}
\affiliation{
  \institution{Université Toulouse UT2J, IRIT, CNRS}
  \city{Toulouse}
  \country{France}
}
\email{andre.peninou@irit.fr}

\author{Olivier Teste}
\affiliation{
  \institution{Université Toulouse UT2J, IRIT, CNRS}
  \city{Toulouse}
  \country{France}
}
\email{olivier.teste@irit.fr}

\begin{abstract}
Federated Learning (FL) is a decentralized paradigm that enables a client-server architecture to collaboratively train a global Artificial Intelligence model without sharing raw data, thereby preserving privacy. A key challenge in FL is Non-IID data. Quantity Skew (QS) is a particular problem of Non-IID, where clients hold highly heterogeneous data volumes. Clustered Federated Learning (CFL) is an emergent variant of FL that presents a promising solution to Non-IID problem. It improves models' performance by grouping clients with similar data distributions into clusters. CFL methods generally fall into two operating strategies. In the first strategy, clients select the cluster that minimizes the local training loss. In the second strategy, the server groups clients based on local model similarities. However, most CFL methods lack systematic evaluation under QS but present significant challenges because of it.

In this paper, we present two main contributions. The first one is an evaluation of state-of-the-art CFL algorithms under various Non-IID settings, applying multiple QS scenarios to assess their robustness. Our second contribution is a novel iterative CFL algorithm, named CORNFLQS, which proposes an optimal coordination between both operating strategies of CFL. Our approach is robust against the different variations of QS settings. We conducted intensive experiments on six image classification datasets, resulting in 270 Non-IID configurations. The results show that CORNFLQS achieves the highest average ranking in both accuracy and clustering quality, as well as strong robustness to QS perturbations. Overall, our approach outperforms actual CFL algorithms.
\end{abstract}


\newcommand\blfootnote[1]{%
  \begingroup
  \renewcommand\thefootnote{}%
  \footnotetext{#1}%
  \addtocounter{footnote}{-1}%
  \endgroup
}

\maketitle

\blfootnote{This is the author’s preprint version of the paper accepted at CIKM 2025.
The final version is published by ACM at \url{https://doi.org/10.1145/3746252.3761216}.}
\section{Introduction}
\label{sec:introduction}

\textbf{Federated Learning (FL)} is a popular privacy preserving paradigm in machine learning that enable learning across distributed devices \cite{FedAvg}. Unlike traditional centralized training, FL allows clients (nodes participating in training) to retain their raw data locally, sharing only the parameters of the locally trained models with a central server to form a global model \cite{FedAvg}. However, a fundamental challenge in FL is \textbf{Non-IID data} (Non-Independent and Identically Distributed data) between clients, leading to sub-optimal model performance \citep{noniid,noniidsurvey}.

FedAvg \cite{FedAvg}, the foundational approach, considered Non-IID data on image classification tasks, but only in limited scenarios involving unbalanced label distributions across clients. Further research \citep{noniid,noniidsurvey,taxonomy} showed that Non-IID data make local models weights drift apart which makes aggregation more complicated. To address this issue, \textbf{FedProx} \cite{fedprox} introduced a proximal term to penalize local updates that deviate excessively from the global model, thereby reducing model drift during training. However, Non-IID data remained problematic, especially when the distribution of the data was very heterogeneous. \textbf{Clustered Federated Learning (CFL)} has emerged as a popular approach to address Non-IID data by grouping clients with similar data distributions into clusters using local models similarity, with each cluster being trained with a dedicated model \citep{cfl,FL+HC,fedprox,ifca,srfca}. By training separate models for each group of clients, CFL reduces model weight drift and improves overall performance. CFL solutions can be categorized in two main paradigms: 
either clients independently choose the cluster that minimizes their local training loss (loss-based), or the server assigns clients to clusters based on similarities between their local models (weight-based).

While CFL methods are typically benchmarked on a data heterogeneity taxonomy \cite{taxonomy}, they often overlook Quantity Skew (QS), a common FL scenario in which clients hold vastly different numbers of samples. This disparity can severely impact the performance of clustering algorithms in CFL, as clients with highly variable dataset sizes may produce divergent model updates that distort clustering metrics and lead to inaccurate groupings, even when they share similar data distributions. 

\begin{figure}[ht]
    \centering
    \hfill
    \begin{minipage}[b]{0.23\textwidth}
        \centering
        \includegraphics[width=\textwidth]{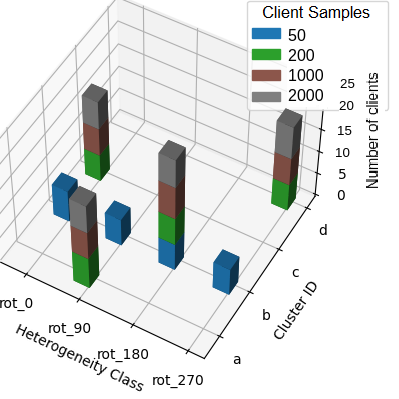}
        \Description{Diagram of a federated learning setup where clients with the same image rotations form groups with quantity skew, labeled QS Type 1.}
        \subcaption{A FL setup where all clients with the same images rotations have QS between them (Quantity Skew Type 1)}
    \end{minipage}
        \hfill
    \begin{minipage}[b]{0.23\textwidth}
        \centering
        \includegraphics[width=\textwidth]{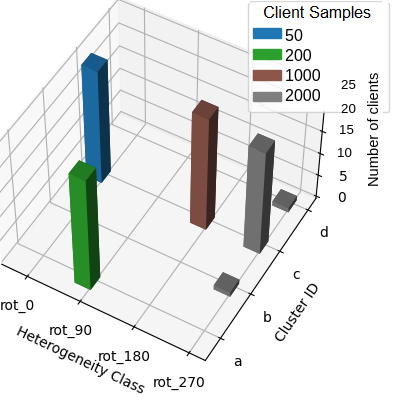}
        \Description{Diagram of a federated learning setup where 100 clients with different image rotations show quantity skew, labeled QS Type 2.}
        \subcaption{Another FL setup where 100 clients with different images rotations have QS between them (Quantity Skew Type 2)}
    \end{minipage}
    \caption{Illustration of CFL experiments with 100 clients on MNIST under concept shift on features with Quantity Skew. Each bar shows the number of clients in a cluster for a given rotation, while the stacked colors indicate their dataset sizes of those clients (see legend for sample size).}
    \label{fig:CFL-QS-exemple}

\end{figure}

To illustrate the problem, consider a standard benchmark setup for image classification in CFL \citep{ifca,srfca}. In this setup, clients are divided into four groups, each with a distinct image rotation (0°, 90°, 180°, or 270°) to simulate data heterogeneity. This type of heterogeneity, where the distribution of input features differs while labels remain consistent, is commonly referred to as concept shift on features. For instance, in a setup configuration with 100 clients, each possessing 500 images, clients sharing the same rotation are all grouped into the same clusters, distinct from cluster with other rotations, using the CFL algorithm (a result consistently confirmed in CFL literature \cite{ifca,cfl,FL+HC,fedgroup}). However, This behavior changes in the two QS scenarios shown in Figure \ref{fig:CFL-QS-exemple}, where client dataset sizes vary significantly. In these cases, CFL fails to correctly group clients by rotation, exposing a key limitation of current methods.

It is important to distinguish QS from class imbalance in centralized or FL settings, which concerns disparities in label distributions. QS, by contrast, specifically refers to differences in the total number of samples held by each client in distributed datasets, regardless of the local or global label distribution. To our knowledge, no prior work systematically studies QS in CFL despite its real-world applications. 

In this paper, our first contribution (Section \ref{sec:CFLQS}) is a comparative evaluation of FedAvg, FedProx, CFL \cite{cfl}, FL+HC \cite{FL+HC}, FedGroup \cite{fedgroup}, IFCA \cite{ifca}, and SRFCA \cite{srfca}, focusing on their behavior under QS and non-QS scenarios. We evaluate these algorithms across six well known image classification datasets: MNIST, Fashion-MNIST, KMNIST, CIFAR-10 and two datasets from MedMNIST dataset collection \cite{medmnist} (OctMNIST and TissueMNIST). Each dataset is declined using three types of Non-IID FL setups from the proposed data heterogeneity taxonomy \cite{taxonomy}, each combined with three QS configurations: one baseline without QS and two representing different QS conditions. The QS setups include one where clients sharing the same data distribution have unequal sample sizes (Quantity Skew Type 1), and another where sample size disparity occurs only between clients with different data distributions (Quantity Skew Type 2).
To ensure reliability, every experiment is repeated across five different random samplings of client data partitions. In total, we evaluate 270 unique FL scenarios for each algorithm. This extensive experimental protocol offers strong empirical evidence of the limitations that current CFL methods face when dealing with QS, in contrast to their more stable performance under non-QS conditions.

Our second contribution is CORNFLQS, a novel CFL algorithm designed for robust performance in both QS and non-QS scenarios. Building on our earlier analysis, CORNFLQS refines clustering by alternating between the two, seeking a common clustering agreement rather than committing to one approach only. This results in improved robustness to QS without compromising performance in non-QS settings compared to state-of-the-art CFL algorithms.

Evaluated on the same 270 FL setups, CORNFLQS shows the highest robustness across all configurations, based on average accuracy ranking, ARI ranking, and variance in local client accuracy. \textbf{The repository including CORNFLQS and all experimental results is available at: \url{https://gitlab.irit.fr/sig/theses/michael-ben-ali/CORNFLQS}.}

The remainder of this paper is organized as follows. In section \ref{sec:relatedwork} we review FL challenges, CFL and data heterogeneity taxonomies. In section \ref{sec:CFLQS}, we offer our first contribution that is the comparative evaluation of QS vs non-QS in CFL. In section \ref{sec:cornflqs-algo}, we explain the choices and mechanisms of CORNFLQS algorithm and finally in section \ref{sec:cornflqs-eval}, we compare its performance to other FL and CFL algorithms.

\section{CFL for Non-IID Data}\label{sec:relatedwork}
Federated Learning (FL) is a decentralized learning paradigm that enables multiple clients to collaboratively train a global model without sharing raw data, thereby preserving privacy and reducing communication costs \cite{FedAvg}. 
In the standard Federated Learning (FL) protocol, known as FedAvg \cite{FedAvg}, the server initializes a model and shares it with all participating clients. Each client trains this model on its private data and then sends the updated local model back to the server. The server aggregates local client models' weights by sample-size based averaging (i.e. proportional to the local dataset sample-size of each client). The resulting global model is sent back to the clients. This cycle, known as a communication round, is a key hyperparameter in all FL algorithms and is typically repeated until a stopping criterion is met. 

In Non-IID data, where client's data is highly heterogeneous, local models weights may drift apart, a phenomenon known as client drift \cite{noniid, noniidsurvey}. This drift can slowly degrade the performance of the global model as the rounds progress. To address this issue, FedProx \cite{fedprox} introduces a proximal term to keep updates close, improving stability and performances. However, FedProx alone cannot fully address the strong heterogeneity introduced in Figure~\ref{fig:CFL-QS-exemple}.

Clustered Federated Learning (CFL) addresses Non-IID data issue by grouping clients with similar data distributions into clusters \cite{cfl}. 
Each cluster trains its own separated FL model, with clients collaborating within the cluster to learn a model corresponding to their shared data distribution.

As discussed in the introduction, Clustered Federated Learning (CFL) methods can be broadly categorized into two main paradigms. \textbf{Weight-based CFL}, such as the original \textbf{CFL} algorithm \cite{cfl}, \textbf{FL+HC} \cite{FL+HC}, or \textbf{FedGroup} \cite{fedgroup}, relies on the weights of locally trained models to measure similarities between clients and applies standard clustering techniques with specific metrics to assign clients to clusters. For example, CFL uses one-shot K-means clustering applied after a determined number of communication rounds, FL+HC applies agglomerative hierarchical clustering with Ward linkage and Euclidean distance similarly, and FedGroup combines the EDC (Euclidean distance of cosine dissimilarity) metric \cite{fedgroup} with K-means clustering following a cold start initialization. \textbf{Loss-based CFL}, including \textbf{IFCA} \cite{ifca} and \textbf{SRFCA} \cite{srfca}, uses loss-based clustering, where multiple cluster-representative models are sent to clients, which iteratively evaluate their local loss to determine the best-fitting model. In practice, IFCA may need to run with several random initializations, selecting the model with the best performance. In SRFCA, all clients cross-evaluate each cluster-representative model on other clients’ local data, and the algorithm applies a metric threshold on the average loss to iteratively merge or split clusters when losses exceed the threshold. The CFL paradigm has shown notable improvements in model effectiveness~\cite{cfl,FL+HC,fedgroup,ifca,srfca}, particularly for highly heterogeneous Non-IID data, which is further detailed in the taxonomy presented in Section~\ref{sec:taxonomy}.

\subsection{Taxonomy of Data Heterogeneity in Federated Learning}\label{sec:taxonomy}
\begin{figure}[ht]
    \centering
    \begin{minipage}{0.15\textwidth}
        \centering
        \includegraphics[width=0.8\textwidth, height=0.7\textwidth]{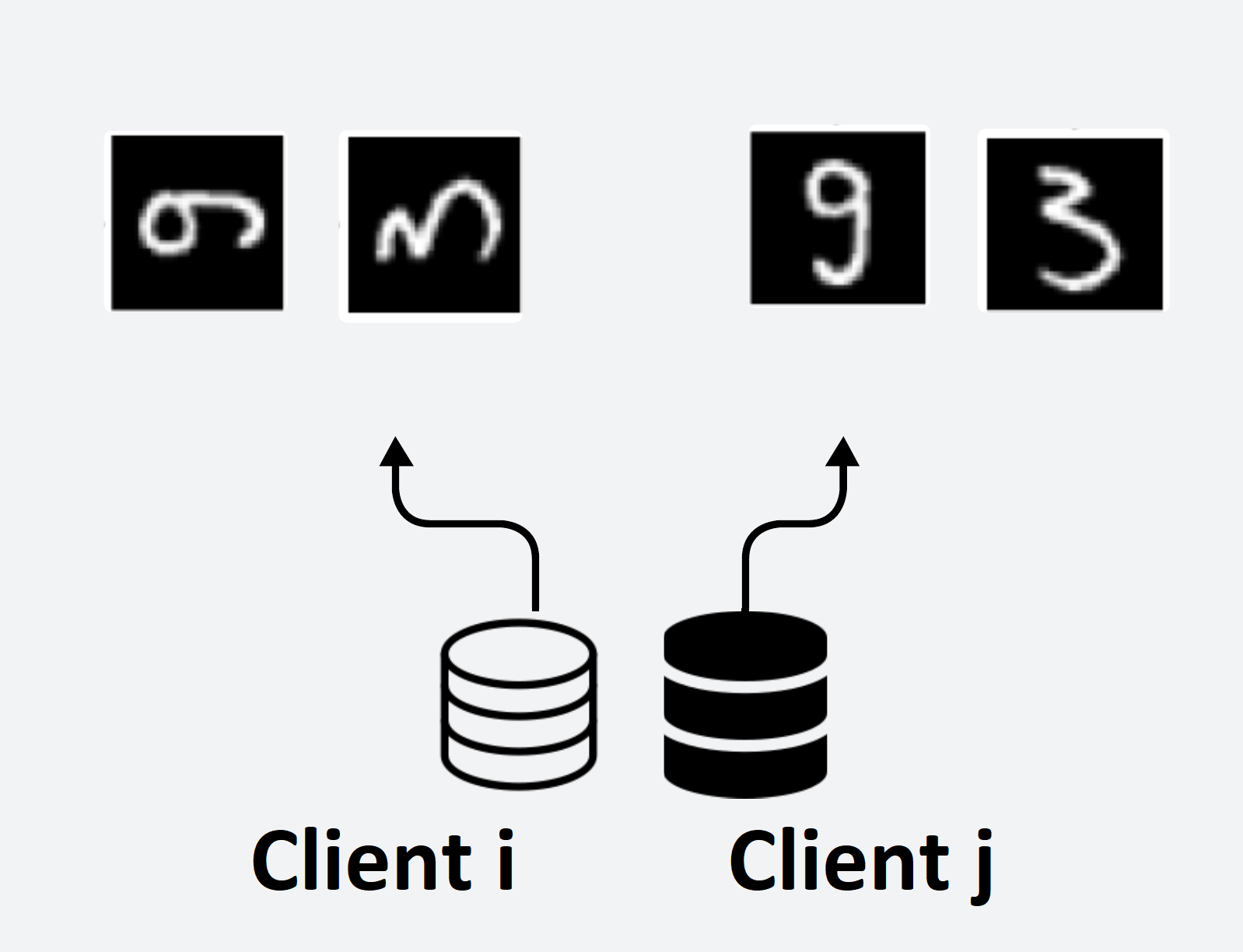}
        \Description{}
        \\ a) Concept Shift on Features
    \end{minipage}
    \hfill
    \begin{minipage}{0.15\textwidth}
        \centering
        \includegraphics[width=0.8\textwidth, height=0.7\textwidth]{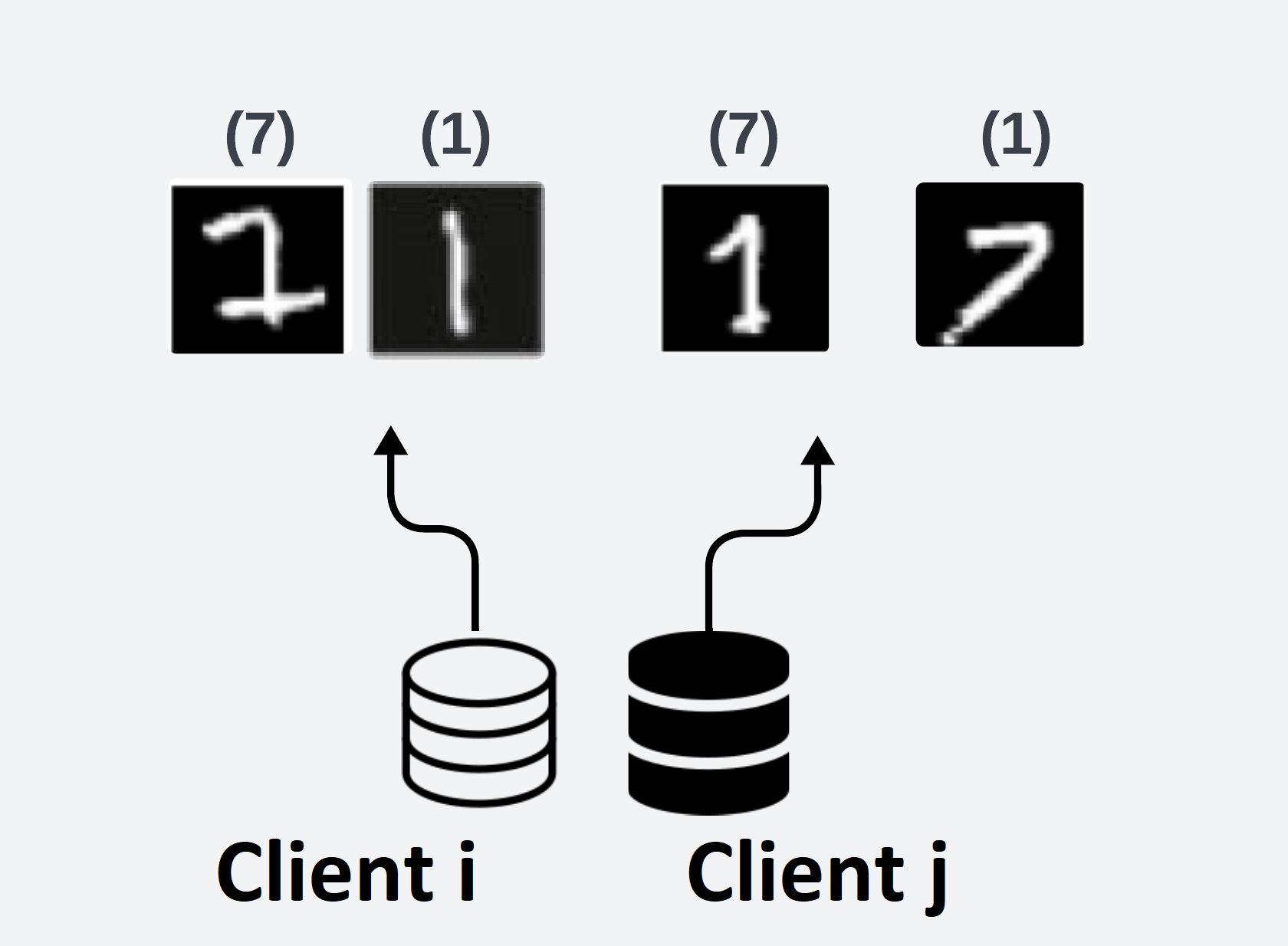}
        \Description{}
        \\ b) Concept Shift on Labels
    \end{minipage}
    \hfill
    \begin{minipage}{0.15\textwidth}
        \centering
        
        \includegraphics[width=0.8\textwidth, height=0.7\textwidth]{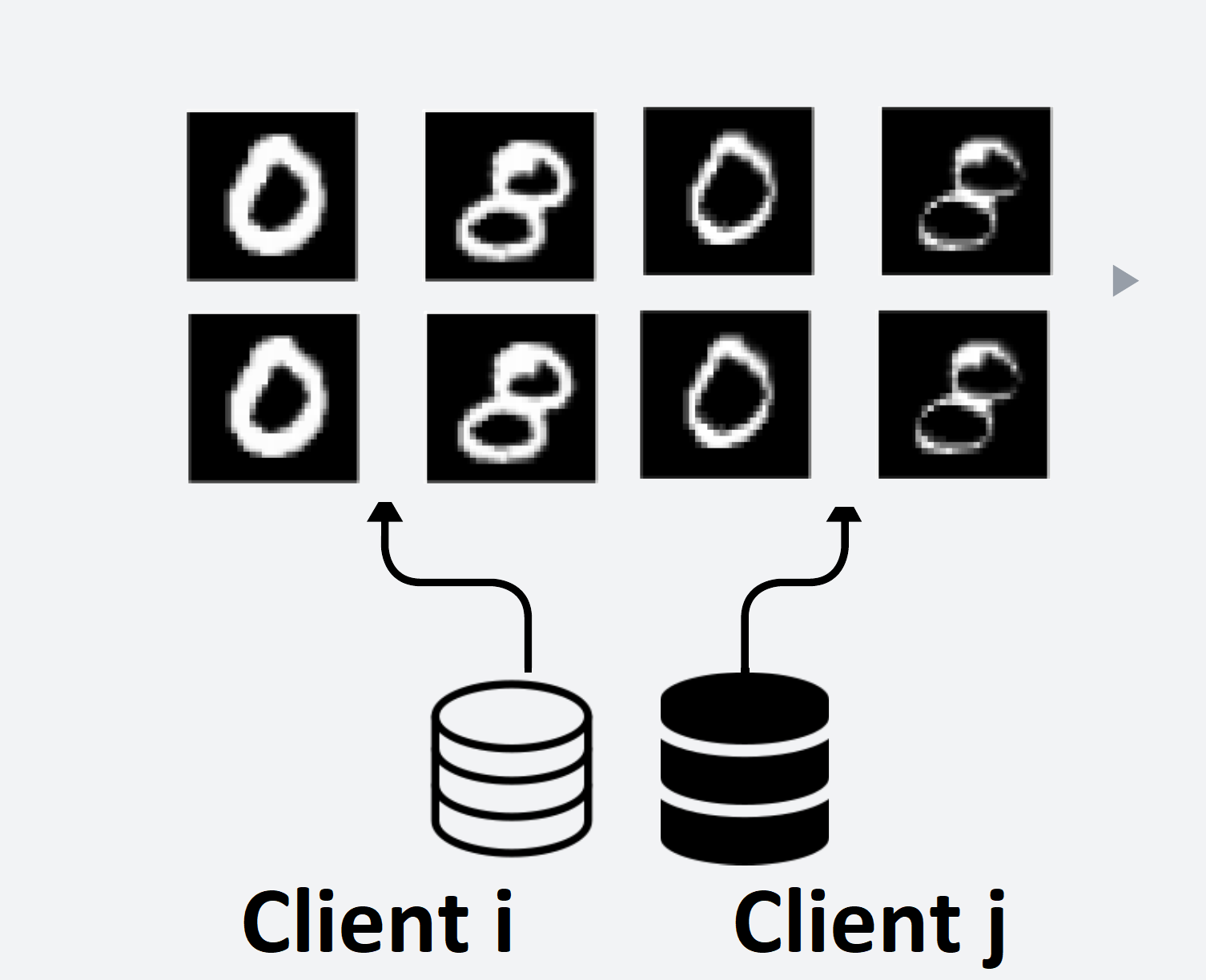}
        \Description{}
        \\ c) Features Distribution Skew
    \end{minipage}
    \caption{Illustration of Non-IID categories for two clients $i$ and $j$ with samples from the MNIST dataset.}
    \label{fig:non_iid_categories}
\end{figure}

Building upon the challenges introduced by Non-IID data in FL, it is crucial to understand the different types of heterogeneity that arise in federated settings \cite{taxonomy}. This section provides an overview of key types of data heterogeneity between two clients illustrated with examples from the MNIST dataset: 
\begin{enumerate}[label=\alph*), left=0pt, itemsep=0.5ex]
\item \textbf{Concept shift on features.} The same labelled data is represented differently in clients' datasets (Figure~\ref{fig:non_iid_categories}.a).
\item \textbf{Concept shift on labels.} Clients' data with similar features are labeled differently (Figure~\ref{fig:non_iid_categories}.b).
\item \textbf{Feature distribution skew.} Clients' data have similar features concept, but with skewed values (Figure~\ref{fig:non_iid_categories}.c).
\end{enumerate}

While the taxonomy of data heterogeneity highlights the diverse and complex nature of Non-IID data in FL, certain groups of clients may exhibit similar patterns in their datasets. These patterns can be captured by local models, which is the intuition behind the concept of CFL.

\subsection{Non-IID Data with QS in CFL}\label{sec:QS}

\textbf{Quantity Skew (QS) }refers to FL setups where clients have highly variable dataset sizes.  The concept of \textbf{heterogeneity classes} is central to understanding the impact of data heterogeneity with QS in CFL. In a Non-IID FL setup with multiple heterogeneous data distributions, clients belong to the same heterogeneity class if they share an independent and identical distribution (IID) within the class, but not with clients outside of it. This classification ensures that, while data distributions remain homogeneous within a class, they exhibit the specified type of heterogeneity when compared to other classes. For example, if a subset of IID clients experiences concept shift on features with other clients, they form a distinct heterogeneity class, although minor empirical variations may still exist within it. Therefore, an effective CFL algorithm should be capable of grouping together clients of the same class in a single cluster; which is the classical CFL assumption.

 We define two fundamental types of QS scenarios, each of which has distinct implications for the federated learning process: 
 \begin{itemize}[label=\textbullet, left=0pt, itemsep=0.5ex]
\item In \textbf{QS-type-1}, variations in client dataset sizes occur within a heterogeneity class. We hypothesize that this may affect how models represent weights for similar data distributions. \textit{Example: Hospitals treating different rare diseases with varying patient volumes within the same disease class.}
\item In \textbf{QS-type-2}, there are disparities in dataset size between different heterogeneity classes. More specifically, the total number of samples of all clients in a given heterogeneity class can vary significantly between classes. This type of QS introduces an additional layer of challenge in FL. The initial global model is biased towards the class with the most samples. \textit{Example: Smartphone users of different age groups generating drastically different data amounts and usage patterns across groups.}
\end{itemize}

Despite extensive research into CFL, the impact of QS has been largely overlooked. The following section uses an experimental setup to highlight the challenges it presents.

\section{Comparative Evaluation of CFL Methods in QS Scenarios}\label{sec:CFLQS}

Image classification is the most widely studied and adopted task for benchmarking CFL methods~\citep{cfl,FL+HC,fedgroup,ifca,srfca}, mainly due to the ease with which distinct data distributions can be created using the data heterogeneity taxonomy~\cite{taxonomy}. Consequently, it is one of the fairest tasks for evaluating CFL algorithms, which is why we have chosen it as the main focus of our evaluation. 
To ensure a diverse and representative evaluation, we select six datasets—MNIST, Fashion-MNIST, KMNIST~\cite{kmnist}, two MEDMNIST~\cite{medmnist} datasets, TissueMNIST, and OctMNIST (all grayscale), along with CIFAR-10 (three-channel images). Each dataset is evaluated independently on three different data heterogeneity setups, as described in Section~\ref{sec:taxonomy} of the taxonomy. For fair comparison, each setup is evenly divided into four heterogeneity classes.

For \textbf{(a) concept shift on features}, heterogeneity in MNIST, Fashion-MNIST, KMNIST~\cite{kmnist}, and CIFAR-10 is introduced by rotating all images of a dataset, as described in the explanation of Figure~\ref{fig:CFL-QS-exemple}. For the MEDMNIST datasets, due to the specific nature of medical images, we apply grayscale inversion and zooming for TissueMNIST and OctMNIST. For \textbf{(b) concept shift on labels}, we use multiple label swaps. For \textbf{(c) feature distribution shift}, we use image dilation and erosion as an alternative to create the necessary Non-IID conditions. In every setup, clients have locally similar and evenly distributed samples per label.

Each setup includes three QS scenarios—non-QS, QS-Type-1, and QS-Type-2—each repeated five times with random samples for 100 clients to ensure robustness. In non-QS, every client has 50 samples per label. In QS1 and QS2, clients are split into four groups with 5, 20, 100, or 200 samples per label. QS1 evenly distributes these groups within each heterogeneity class, while QS2 assigns only one group per class. Each client has a local test set, sampled from the dataset’s test set, with the same test set shared by all clients within a given heterogeneity class and distributed evenly across classes. Altogether, these configurations total 270 experiments for thorough evaluation.

\subsection{Federated Learning Setup}\label{sec:setup}
The architecture used for training our FL models varies depending on the dataset. For MNIST, Fashion-MNIST, and KMNIST, a fully connected neural network with 200 neurons in the hidden layer is employed. For OctMNIST and TissueMNIST, a convolutional neural network (CNN) with two convolutional layers, max pooling, and one fully connected layer is used. Finally, for CIFAR-10, we use a CNN with four convolutional layers and two residual blocks, and apply standard data augmentation methods such as random horizontal flip and random crop to ensure sufficient training data during training. The choice of these architectures is inspired by CFL experimentation from the literature \cite{ifca, cfl, FL+HC, fedgroup, srfca}.

\sloppy{
The federated learning algorithms evaluated in this section (which were introduced in Section~\ref{sec:relatedwork}) include CFL and FL+HC, which apply clustering after half of the communication rounds; FedGroup, which uses clustering following a cold-start initialization; IFCA, where five random initializations are run and the best-performing model is selected; and SRFCA, which identifies clusters using a metric threshold.}

All algorithms are evaluated using similar hyperparameters. The number of clusters is set to four for all experiments, corresponding to the number of heterogeneity classes. The only exception is SRFCA, which performs a grid search for the metric threshold, selecting values between the first decile and the first quartile of the observed metric distribution to determine the number of clusters. Communication rounds are fixed at 20 with 10 local epochs for fully connected models, while CNN-based models run for 100 rounds with 5 local epochs per round.

\subsection{Impact of Quantity Skew on CFL algorithms}

The aim of this section is to demonstrate that our experimental setup effectively highlights the challenges posed by QS in CFL. In Tables~\ref{tab:non-QS} (non-QS), \ref{tab:qs1} (QS1), and \ref{tab:qs2} (QS2), we report three metrics. Each reported value is accompanied by its corresponding standard deviation across experimental runs. In all tables, the best value for each metric is highlighted in bold, and the second-best is underlined. The considered metrics are : 

\begin{itemize}[label=\textbullet, left=0pt, itemsep=0.5ex]
\item The mean global model accuracy (\textbf{Acc}). It is calculated as the average accuracy of each cluster model on its assigned clients' test sets. 

\item In a particular setup, the standard deviation of client-level accuracies reflects the performance disparity across clients within that setup. A high value typically indicates that clients with heterogeneous data distributions are grouped together, leading to biased performance—often a sign of poor clustering and low ARI. The reported values here, noted as\textbf{ Client Acc Std}, are the mean and standard deviation of the standard deviations of client-level accuracies, computed across different experimental runs.

\item The mean Adjusted Rand Index (\textbf{ARI}). We also use the Adjusted Rand Index (ARI), which measures agreement between predicted clusters and the heterogeneity classes defined in Section \ref{sec:QS}. An ARI near 1 means near-perfect clustering, while around 0 indicates random assignment.
\end{itemize}

Our objective with CFL is to maximize global accuracy while minimizing the standard deviation in client-level accuracy. The ARI is not merely a supporting metric but a key indicator of model performance in CFL, i.e., a high ARI implies that clients are clustered correctly according to their data distributions, which directly contributes to improved local and global accuracy. Thus, the higher the ARI, the more effective the clustering strategy and the better the overall performance. Ideally, a robust algorithm should consistently achieve high scores across all metrics and scenarios. 

In the non-QS setting (Table \ref{tab:non-QS}), effectiveness across algorithms is relatively comparable. IFCA, followed closely by FedGroup and then FL+HC, shows marginally better accuracy. However, these results do not account for the ranking of each algorithm across individual experiments, which we explore further in Section \ref{sec:cornflqs-eval}. In the QS1 and QS2 scenarios, IFCA (Table \ref{tab:qs1}) and FL+HC (Table \ref{tab:qs2}) outperform other methods on average. At first glance, IFCA appears to be the most robust in the face of QS variations. However, this is offset by the fact that it relies on randomness and requires multiple runs to achieve optimal performance. By contrast, deterministic methods such as FL+HC can achieve better results with a single run with both non-QS and QS2 settings.

SRFCA tends to underperform globally, which can be attributed to significantly lower ARI scores. As SRFCA does not consider the number of clusters as a hyperparameter, it often struggles to identify the correct clustering configuration, relying solely on a threshold-based metric. Similarly, FedGroup was originally evaluated on a large number of clients with fewer data samples per client. This setting does not translate well to QS scenarios and may explain its underwhelming performance here.

\begin{table}[ht]
\centering
\tiny
\begin{subtable}[t]{0.48\textwidth}
\centering
\resizebox{0.8\textwidth}{!}{%
\begin{tabular}{|l|c|c|c|}
\hline
\textbf{Algorithm} & \textbf{Acc} & \textbf{Client Acc Std} & \textbf{ARI} \\
\hline
cfl       & 71.00 $\pm$ 15.71 & 7.13 $\pm$ 7.76 & 0.68 $\pm$ 0.37 \\
fl+hc      & 71.60 $\pm$ 15.24 & 5.76 $\pm$ 7.59 & \underline{0.84} $\pm$ 0.26 \\
fedgroup  & \underline{71.71} $\pm$ 15.16 & \textbf{5.42} $\pm$ 4.75 & 0.69 $\pm$ 0.36 \\
ifca      & \textbf{72.00} $\pm$ 15.07 & \underline{6.15} $\pm$ 1.78 & \textbf{0.85} $\pm$ 0.20 \\
srfca     & 62.47 $\pm$ 20.26 & 6.07 $\pm$ 6.51 & 0.50 $\pm$ 0.46 \\
\hline
\end{tabular}
}
\caption{non-QS setting.}
\label{tab:non-QS}
\end{subtable}

\begin{subtable}[t]{0.48\textwidth}
\centering
\resizebox{0.8\textwidth}{!}{%
\begin{tabular}{|l|c|c|c|}
\hline
\textbf{Algorithm} & \textbf{Acc} & \textbf{Client Acc Std} & \textbf{ARI} \\
\hline
cfl       & 67.29 $\pm$ 13.77 & 12.33 $\pm$ 9.80 & 0.37 $\pm$ 0.35 \\
fl+hc      & \underline{68.00} $\pm$ 13.80 & 13.57 $\pm$ 11.30 & 0.43 $\pm$ 0.39 \\
fedgroup  & 67.64 $\pm$ 15.95 & 9.32 $\pm$ 7.65 & 0.42 $\pm$ 0.41 \\
ifca      & \textbf{73.18} $\pm$ 14.48 & \underline{6.94} $\pm$ 3.22 & \textbf{0.82} $\pm$ 0.23 \\
srfca     & 65.65 $\pm$ 19.53 & \textbf{4.95} $\pm$ 4.44 & \underline{0.63} $\pm$ 0.37 \\
\hline
\end{tabular}
}
\caption{QS1 setting.}
\label{tab:qs1}
\end{subtable}

\begin{subtable}[t]{0.48\textwidth}
\centering
\resizebox{0.8\textwidth}{!}{%
\begin{tabular}{|l|c|c|c|}
\hline
\textbf{Algorithm} & \textbf{Acc} & \textbf{Client Acc Std} & \textbf{ARI} \\
\hline
cfl       & 63.99 $\pm$ 9.34  & 18.39 $\pm$ 10.01 & 0.38 $\pm$ 0.28 \\
fl+hc      & \textbf{74.08} $\pm$ 10.84 & \underline{8.26} $\pm$ 8.49 & \textbf{0.82} $\pm$ 0.32 \\
fedgroup  & 69.75 $\pm$ 14.03 & 9.22 $\pm$ 7.04 & 0.64 $\pm$ 0.39 \\
ifca      & \underline{71.94} $\pm$ 11.09 & 10.03 $\pm$ 4.22 & \underline{0.76} $\pm$ 0.22 \\
srfca     & 65.31 $\pm$ 18.85 & \textbf{7.22} $\pm$ 3.92 & 0.59 $\pm$ 0.36 \\
\hline
\end{tabular}
}
\caption{QS2 setting.}
\label{tab:qs2}
\end{subtable}

\caption{Average Performance comparison of CFL algorithm in different heterogeneity settings with non-QS, QS1 and QS2 over all datasets described in Section \ref{sec:CFLQS}.}
\label{tab:perf_comparison}
\end{table}

\begin{figure}[ht]
\includegraphics[width=0.65\linewidth]{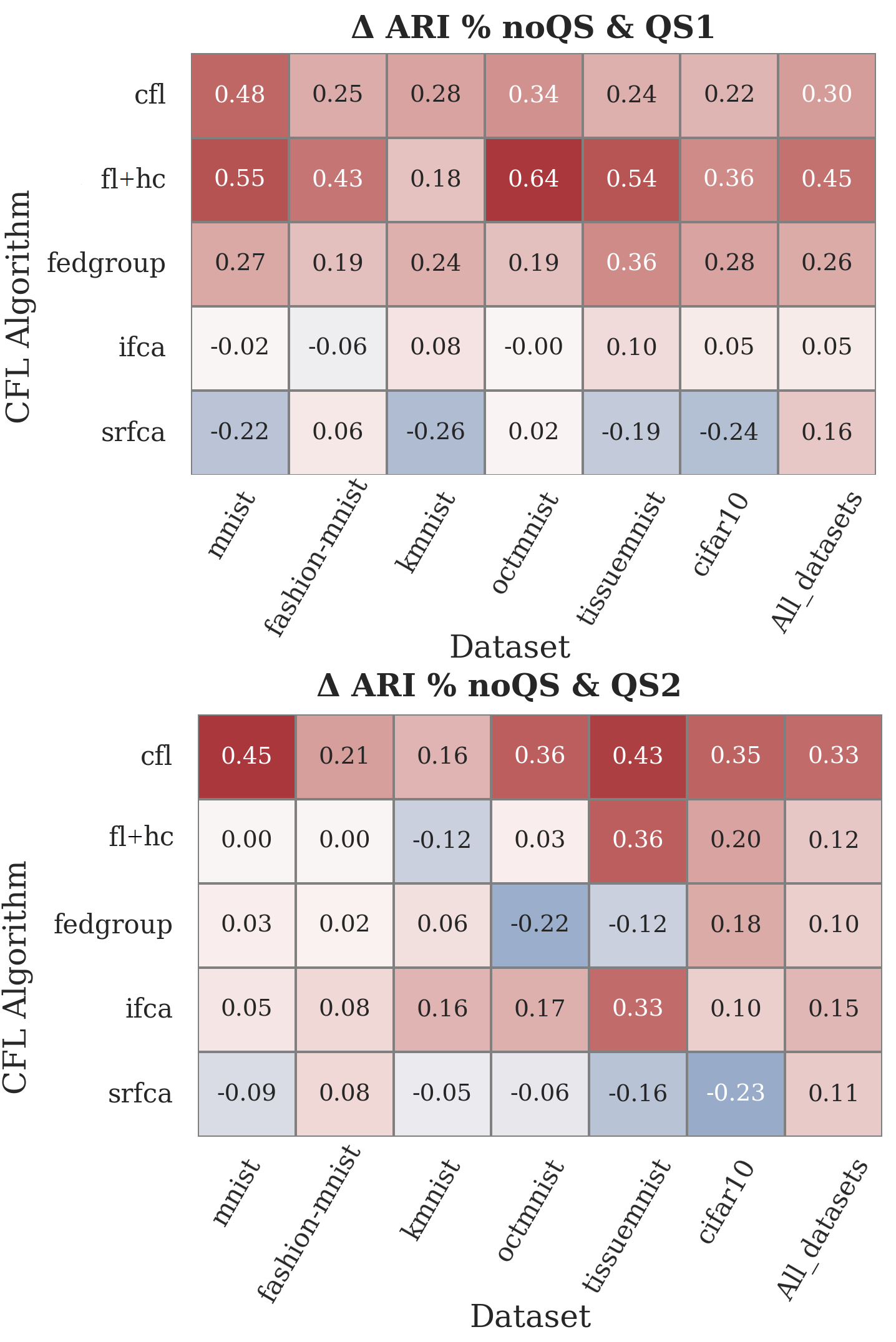}
\Description{}
\caption{$\Delta$ ARI Heatmaps of CFL algorithms between non-QS and QS setups.}
\label{fig:delta_metrics}
\end{figure}

To better understand the effect of QS on CFL algorithms, we analyze the discrepancies in behavior according to QS and non-QS setups. Figure~\ref{fig:delta_metrics} shows two heatmaps visualizing the average differences in ARI — here interpreted as the percentage of clients correctly grouped with other clients from the same heterogeneity class — between aligned QS and non-QS configurations (denoted as $\Delta$ between non-QS and QS1 or QS2). These values, averaged over all configurations introduced in Section~\ref{sec:CFLQS} and based on the same experiments as those reported in Table~\ref{tab:perf_comparison}, quantify the impact of QS on clustering quality.
 The heatmaps show how strong the variations are: red tones indicate clustering (ARI) degradations due to QS, while blue tones show improvement. The y-axis lists the algorithms, the x-axis lists the datasets, and an additional column  shows the average results across all datasets. As a robust algorithm should minimize these discrepancies regardless of direction, the average column reports the mean absolute value. These visualizations demonstrate how QS can significantly impact the reliability of clustering and, consequently, CFL performance.

We focus on $\Delta$ ARI and the change in the standard deviation of client accuracy, as they offer clearer and more consistent insights into the effects of QS. Looking at Figure~\ref{fig:delta_metrics} QS1 scenario (top of Figure~\ref{fig:delta_metrics}), we observe that IFCA exhibits the greatest stability in terms of clustering quality, whereas weight-based clustering methods (cfl, FL+HC and fedgroup) show notable discrepancies. Interestingly, both QS1 and QS2 appears to improve the clustering performance of SRFCA. This can be attributed to the fact that, in a QS setup, when two clients of the same heterogeneity class exchange models, the client with more data improves its counterpart's model's performance. Furthermore, since SRFCA uses trimmed mean aggregation, the influence of models trained on small data quantities of data have less influence, which improves clustering quality in this scenario.

For the QS2 scenario (bottom of Figure~\ref{fig:delta_metrics}), the results vary slightly across datasets. FL+HC consistently demonstrates the best stability, confirming the trends observed in Table~\ref{tab:perf_comparison}.

Although $\Delta$ accuracy may initially seem like a more intuitive metric for assessing performance shifts than the ARI metric, its interpretation is often misleading. This is particularly relevant in situations where QS changes the composition of clusters in a way that has a disproportionate impact on clients with small sample sizes. For this reason, we have deliberately excluded $\Delta$ accuracy as a general evaluation metric. However, we highlight specific cases to illustrate how QS can impact differences in client accuracies. 

In the QS1 scenario, when analyzing results at the granularity of a single run rather than aggregated outcomes, we find that average accuracy differences can obscure crucial effects of misclustering. Table~\ref{tab:toy_qs1_example} illustrates this with MNIST and CIFAR-10 under concept-shift-on-features.
For MNIST in the non-QS setting, where the Adjusted Rand Index (ARI) is $1.0$, CFL reaches an average accuracy of $83.34\%$, while FedAvg lags behind at $55.90\%$. This illustrates the drawback of naïvely aggregating across heterogeneous clients. In the QS1 setting, however, the situation becomes more nuanced: while FedAvg achieves only $50.52\%$, CFL improves to $77.69 \pm 29.25\%$. Yet, this apparent improvement conceals a severe disparity: clients with high sample-size datasets (HS, $200$ samples per label) achieve $87.82\%$, whereas low-sample clients (LS, $5$ samples per label) collapse to just $22.21\%$. The cluster-wide average is thus dominated by HS clients, masking the fact that LS clients experience significant degradation. This highlights that $\Delta$ accuracy is an unreliable indicator of QS effectiveness, since it does not capture the distribution of performance across clients.
CIFAR-10 follows a similar but less severe pattern. Without QS, FedAvg achieves $65.91\%$ while CFL reaches $72.33 \pm 8.34\%$ with a low ARI of $0.25$. Under QS1, FedAvg rises to $73.94\%$, consistent with the intuition that larger effective data volume benefits local CNN training \cite{cnnlearning}. CFL also improves to $78.53 \pm 11.57\%$, but again the gains are not uniform: HS clients achieve $85.00\%$, while LS clients fall back to $53.94\%$. 
Taken together, these toy examples show why ARI provides a more faithful signal than $\Delta$ accuracy for assessing QS: it directly reflects whether clients are grouped according to their heterogeneity classes, while average accuracies can mask severe drops for vulnerable LS clients. This helps explain the broader results reported in Figure~\ref{fig:delta_metrics} and Table~\ref{tab:perf_comparison}.

In the QS2 scenario, clustering behaviors differ notably from QS1 due to the structure of heterogeneity classes. Since QS is applied within each heterogeneity class, all clients in a given class share the same sample size. However, despite this uniformity, loss-based clustering can become less reliable than weight-based methods. This is especially evident in challenging settings like concept shift on labels, where clients have partially swapped labels. In such cases, low-sample-size clients can still benefit from the model updates of high-sample-size clients in other heterogeneity classes that more effectively learn the shared, non-swapped labels. As a result, even though these clients belong to different classes, their models may exhibit lower loss on each other’s data. 

\begin{table}[ht]
\footnotesize
\centering
\caption{Illustrative toy example of accuracy (\%) for FedAvg and CFL under concept-shift-on-features with MNIST and CIFAR-10 with non-QS and QS1. HS = high-sample clients, LS = low-sample clients.}
\label{tab:toy_qs1_example}
\begin{tabular}{lccccc}
\toprule
Dataset - Scenario & \multicolumn{1}{c}{FedAvg} & \multicolumn{4}{c}{CFL}  \\
\cmidrule(lr){2-2} \cmidrule(lr){3-6} 
  & Acc & Acc & ARI & HS Acc & LS Acc \\
\midrule
MNIST non-QS    & 55.90 & 83.34 $\pm$ 3.93 & 1.00 & -- & --   \\
CIFAR-10 non-QS  & 65.91 & 72.33 $\pm$ 8.34 & 0.25 & -- & --  \\ 
MNIST QS1       & 50.52 & 77.69 $\pm$ 29.25 & 0.51 & 87.82 & 22.21 \\ 
CIFAR-10 QS1    & 73.94  & 78.53 $\pm$ 11.57 & 0.24 & 85.00 & 53.94\\ 
\bottomrule
\end{tabular}
\end{table}

This effect can mislead loss-based clustering (IFCA, SRFCA) because the loss no longer reliably reflects intra-class similarity. By contrast, weight-based clustering (CFL, FL+HC, FedGroup) uses model parameters that are shaped by local training. As clients within the same heterogeneity class have identical data quantities and distributional characteristics, their weights tend to evolve in a more similar manner. This makes weight-based similarity generally more robust in this scenario. However, this does not mean that weight-based clustering is immune to the effects of QS2 since model weights can still be influenced by label distribution shifts and subtle representational differences. That said, the distinction is somewhat less pronounced in CNN-based models with data augmentation, where exposure to greater data variation allows even low-sample clients to learn slightly richer representations, partially mitigating the issues caused by small sample sizes.

To conclude, our experiments highlight the challenges introduces by QS in CFL. While some algorithms, like IFCA or FL+HC, demonstrate resilience to specific QS variations, even these face robustness challenges in other scenarios. This highlights the importance of both model weights and model loss in clustering. They offer complementary insights that help to address various QS and CFL conditions. The results from CIFAR-10 point the need for deep analysis of how datasets and data distributions affect the effectiveness of clustering under QS. Given these challenges, we propose CORNFLQS, a robust solution combining the strengths of FL+HC and IFCA. This new approach is specifically designed to better handle the complexities of Non-IID data with QS.

\section{A Robust Solution for Non-IID Data with Quantity Skew}\label{sec:cornflqs-algo}

CORNFLQS (Clustering Optimal Research between Nodes for Federated Learning with Quantity Skew) is a novel approach that combines the strengths of both model-based and loss-based clustering methods to overcome the challenges posed by QS, as observed in Section~\ref{sec:CFLQS}. The main steps of the CORNFLQS algorithm are illustrated in Algorithm~\ref{alg:cornflqs}. It requires $N$ the number of rounds, a set of clients, denoted by $C$, with $|C|$ representing the total number of clients. Each client $c_i \in C$ possesses its own dataset $D_i$ of size $|D_i|$. An important hyperparameter of CORNFLQS is the number of clusters $K$, which determines how many groups of clients are formed during training. 

The algorithm follows four main phases, detailed in Algorithms~\ref{alg:init} to \ref{alg:FedAvg}. It starts with an initialization phase (Algorithm~\ref{alg:init}) that performs an initial federated aggregation using all clients. This is followed by the CORN clustering step (Algorithm~\ref{alg:corn}), combining weight-based and loss-based clustering. A loss-based CFL algorithm (Algorithm~\ref{alg:client-side}) is then applied to stabilize clusters. Finally, standard FedAvg is used inside each cluster for the remaining rounds until reaching $N$. Comments in each sub-algorithm indicate whether operations occur on the server or client side.

\begin{algorithm}[ht]
\footnotesize
\caption{CORNFLQS}
\label{alg:cornflqs}
\begin{algorithmic}
\State \textbf{Input} \mbox{Clients set $C$, Number of Rounds $N$, Number of 
Clusters $K$}   
\end{algorithmic}
\begin{algorithmic}[1]
\State $\textsc{Initialization}$ \Comment{Algorithm 2}
\State $\textsc{CORN}$           \Comment{Algorithm 3}
\State $\textsc{LossBasedCFL}$   \Comment{Algorithm 4}
\State $\textsc{FedAvgforCFL}$   \Comment{Algorithm 5}
\end{algorithmic}
\end{algorithm}

CORNFLQS begins with the initialization phase (Algorithm~\ref{alg:init}). The server initializes a model $W^{(0)}$ and broadcasts it to all clients. Clients train locally and send updated models back, which the server averages and rebroadcast. A second round of local training follows, producing updated models $w_i^{(0)}$ sent back to the server. This process ensures a fair, non-random starting point for the CFL process, avoiding bias toward high samples-size clients in QS scenario. 

\begin{algorithm}[ht]
\footnotesize
\begin{algorithmic}[1]
\caption{$\textsc{Initialization}$}
\label{alg:init}
\State $W^{(0)}$ $\gets$ Initialize model \Comment{Server}
\State Broadcast $W^{(0)}$ to $C$  \Comment{Server}

\ForAll{$c_i \in C$ \textbf{in parallel}}  \Comment{Client}
    \State $w_i^{(0)}$ $\gets$ Train model $W^{(0)}$ locally on dataset $|D_i|$ 
    \State Transfer $w_i^{(0)}$ to Server
\EndFor
\State $W^{(0)}$ $\gets$ $\displaystyle \frac{1}{|C|} \sum_{i=1}^{|C|} w_i^{(0)}$  \Comment{Server}
\State Broadcast $W^{(0)}$ to $C$ \Comment{Server}
\ForAll{$c_i \in C$ \textbf{in parallel}} \Comment{Client}
    \State $w_i^{(0)}$ $\gets$ Train model $W^{(0)}$ locally on dataset $|D_i|$ 
    \State Transfer $w_i^{(0)}$ to Server
\EndFor
\end{algorithmic}
\end{algorithm}

\begin{algorithm}[ht]
\footnotesize
\caption{$\textsc{CORN}$ (Clustering Optimal Research between Nodes)}
\label{alg:corn}
\begin{algorithmic}[1]
\State $r \gets 0$ \Comment{Server}
\Do    
    \State $r \gets r+1$ \Comment{Server}

    \State $\mathcal{K}^{(r)} \gets \textsc{Clustering}(\{w_i^{(r-1)}\}_{i=1}^{|C|}, K)$ \Comment{Server}
    \For{$k \in [1..K]$} \Comment{Server}
        \State $\displaystyle W^{(k)} \gets \frac{1}{\sum_{j \in K^{(r)}_{k}} |D_j|} \sum_{i \in K^{(r)}_{k}} |D_i| \cdot w_i$
    \EndFor
    \State Broadcast $ W = \{ W^{(1)}, \dots,  W^{(K)} \}$ to $C$ \Comment{Server}
    \ForAll{$c_i \in C$ \textbf{in parallel}} \Comment{Client}
        \For{$k \in [1..K]$}
            \State Compute loss $\mathcal{L}_{i,k}(D_i)$ from $W^{(k)}$
        \EndFor
    \State $\displaystyle k_i^* = \textsc{argmin}_{k \in [1..K]} \mathcal{L}_{i,k}(D_i)$
    \State $w_i^{(r)}$ $\gets$ Train model $W^{(k_i^*)}$ locally on dataset $|D_i|$ 
    \State Transfer $w_i^{(r)}$ and $k_i^*$ to Server
    \EndFor
    \State $\mathcal{K}^{(r+1)} \gets \textsc{ClusterAssignment}(w_i^{(r)}, k_i^*)$ \Comment{Server}
\doWhile{$r \leq N/2$ \textbf{and} $K^{(r)} \neq K^{(r+1)}$}
\end{algorithmic}
\end{algorithm}

\begin{algorithm}[ht]
\footnotesize
\caption{$\textsc{LossBasedCFL}$}
\label{alg:client-side}
\begin{algorithmic}[1]

\Do    
    \State $r \gets r+1$ \Comment{Server}
    \For{$k \in [1..K]$} \Comment{Server}
        \State $\displaystyle W^{(k)} \gets \frac{1}{\sum_{j \in K^{(r)}_{k}} |D_j|} \sum_{i \in K^{(r)}_{k}} |D_i| \cdot w_i$
    \EndFor
    \State Broadcast $ W = \{ W^{(1)}, \dots,  W^{(K)} \}$ to $C$ \Comment{Server}
    \ForAll{$c_i \in C$ \textbf{in parallel}} \Comment{Client}
        \For{$k \in [1..K]$}
            \State Compute loss $\mathcal{L}_{i,k}(D_i)$ from $W^{(k)}$
        \EndFor
    \State $\displaystyle k_i^* = \textsc{argmin}_{k \in [1..K]} \mathcal{L}_{i,k}(D_i)$
    \State $w_i^{(r)}$ $\gets$ Train model $W^{(k_i^*)}$ locally on dataset $|D_i|$ 
    \State Transfer $w_i^{(r)}$ and $k_i^*$ to Server
    \EndFor
    \State $\mathcal{K}^{(r+1)} \gets \textsc{ClusterAssignment}(w_i^{(r)}, k_i^*)$ \Comment{Server}
\doWhile{$r \leq N$ \textbf{and} $K^{(r)} \neq K^{(r+1)}$}

\end{algorithmic}
\end{algorithm}

\begin{algorithm}[ht]
\footnotesize
\caption{$\textsc{FedAvgforCFL}$}
\label{alg:FedAvg}
\begin{algorithmic}[1]
\While{$r \leq N$}    
    \State $r \gets r+1$ \Comment{Server}
    \For{$k \in [1..K]$} \Comment{Server}
        \State $\displaystyle W^{(k)} \gets \frac{1}{\sum_{j \in K^{(r)}_{k}} |D_j|} \sum_{i \in K^{(r)}_{k}} |D_i| \cdot w_i$
    \EndFor
    \State $\forall c_i \in C$, Transfer $W^{(k_i^*)}$ to $c_i$  \Comment{Server}
    \ForAll{$c_i \in C$ \textbf{in parallel}} \Comment{Client}
    \State $w_i^{(r)}$ $\gets$ Train model $W^{(k_i^*)}$ locally on dataset $|D_i|$ 
    \State Transfer $w_i^{(r)}$ to Server
    \EndFor
\EndWhile

\end{algorithmic}
\end{algorithm}

Algorithm~\ref{alg:corn} (CORN: Cluster Optimal Research between Nodes) drives the core clustering process by alternating weight-based and loss-based clustering to find an optimal agreement between clients and server. It begins by initializing the round number $r$ to $0$ (line 1). Each iteration of the do-while loop corresponds to a communication round between the server and clients. In each communication round, client model weights $w_i^{(r-1)}$ are clustered using a chosen algorithm (line 4), producing memberships $K^{(r)}$, where $K_k^{(r)}$ denotes the set of indices of clients in cluster $k$. The server then aggregates models within each cluster proportionally to sample size (lines 5--6) and broadcasts the cluster models to all clients (line 8).

Each client computes the loss for each received model $\mathcal{L}_{i,k}(D_i)$ on its local data (lines 10--11), selects the cluster $k_i^*$ with the lowest loss (line 13), train that model to get $w_i^{(r)}$, and sends both the updated model and cluster choice $k_i^*$ to the server (lines 14–15). The server updates cluster memberships to form $K^{(r+1)}$. The process repeats while $r \leq N/2$ and $K^{(r)} \neq K^{(r+1)}$, ensuring effective clustering when both clustering steps agree — an agreement reached before $N/2$ suggests that the clustering is optimal.

The next phase, shown in Algorithm~\ref{alg:client-side}, follows a process similar to CORN but removes line 4 from Algorithm~\ref{alg:corn} and extends the while-loop to run while $r \leq N$.
In this LossBasedCFL subroutine, the server no longer clusters model weights. Instead, it relies solely on selecting cluster based on loss to update cluster memberships on. In each iteration, it computes a new clustering $K^{(r+1)}$ and compares it with the previous round clustering $K^{(r)}$, (the first $K^{(r)}$ is the one obtained at the end of CORN). This process continues either until the clustering remains stable between two consecutive rounds or until the predefined maximum number of rounds is reached ($N$).

\begin{table*}[t]
\centering
\caption{Benchmarking results - Global Ranking across datasets.}
\label{tab:global_results}
\resizebox{\textwidth}{!}{%
\begin{tabular}{ll
cccc cccc cccc cccc cccc cccc cccc}
\toprule
Algorithm
& \multicolumn{3}{c}{All Datasets}
& \multicolumn{3}{c}{MNIST} 
& \multicolumn{3}{c}{Fashion-MNIST} 
& \multicolumn{3}{c}{KMNIST} 
& \multicolumn{3}{c}{CIFAR-10}
& \multicolumn{3}{c}{OctMNIST}
& \multicolumn{3}{c}{TissueMnist}\\
\cmidrule(lr){3-5} \cmidrule(lr){6-8} \cmidrule(lr){9-11} \cmidrule(lr){12-14} \cmidrule(lr){15-17} \cmidrule(lr){18-20} \cmidrule(lr){21-23}
& & Rank & Acc$\pm$Std & ARI 
& Rank & Acc$\pm$Std & ARI 
& Rank & Acc$\pm$Std & ARI 
& Rank & Acc$\pm$Std & ARI 
& Rank & Acc$\pm$Std & ARI 
& Rank & Acc$\pm$Std & ARI 
& Rank & Acc$\pm$Std & ARI \\
\midrule
cornflqs & &\textbf{2.24} & \textbf{73.06$\pm$14.88} & \textbf{0.90} & \textbf{2.15} & \textbf{89.08$\pm$3.09} & \underline{0.96} & \underline{2.31} & \textbf{82.77$\pm$1.31} & \textbf{1.00} & \textbf{2.02} & \textbf{65.86$\pm$2.32} & \textbf{0.96} & \textbf{1.89} & \textbf{76.31$\pm$5.42} & \textbf{0.78} & \underline{3.07} & \underline{65.85$\pm$4.97} & \textbf{0.82}& \textbf{2.33} & \textbf{40.09$\pm$1.52} & \textbf{0.86} \\
fl+hc & & \underline{3.24} & \underline{69.47$\pm$15.86} & 0.68 & 2.98 & 85.23$\pm$8.68 & 0.82 & 3.60 & 80.03$\pm$4.93 & 0.86 & \underline{2.27} & 64.83$\pm$4.65 & 0.86 & \underline{2.73} & \underline{75.94$\pm$5.72} & 0.55 & 3.33 & 64.69$\pm$4.29 & 0.64 & 4.40 & 35.67$\pm$5.77 & \underline{0.44} \\
ifca & & 3.47 & 67.62$\pm$16.63 & \underline{0.75} & \underline{2.80} & \textbf{89.08$\pm$3.09} & \textbf{0.97} & 2.73 & 81.88$\pm$2.52 & 0.93 & 3.04 & \underline{65.16$\pm$3.11} & \underline{0.92} & 6.49 & 62.52$\pm$3.25 & \underline{0.66} & \textbf{2.27} & \textbf{66.56$\pm$3.73} & \textbf{0.69} & \underline{3.49} & 36.21$\pm$6.80 & 0.26 \\
fedgroup & & 4.12 & 67.63$\pm$16.76 & 0.56 & 3.24 & \underline{88.06$\pm$4.18} & 0.90 & \textbf{2.16 }& \underline{82.13$\pm$2.26} & 0.93 & 3.84 & 63.79$\pm$4.77 & 0.79 & 4.44 & 68.30$\pm$6.09 & 0.14 & 6.27 & 56.25$\pm$8.44 & 0.23 & 4.53 & \underline{36.28$\pm$3.48} & 0.40 \\
cfl & & 4.51 & 65.56$\pm$15.75 & 0.46 & 5.07 & 78.58$\pm$9.33 & 0.55 & 4.98 & 74.62$\pm$8.62 & 0.65 & 4.67 & 57.68$\pm$7.55 & 0.49 & 2.93 & 73.91$\pm$6.36 & 0.44 & 4.29 & 63.05$\pm$5.40 & 0.38 & 5.87 & 33.36$\pm$4.65 & 0.24 \\
srfca & & 4.88 & 65.82$\pm$16.26 & 0.53 & 5.04 & 84.74$\pm$7.04 & 0.79 & 5.00 & 78.92$\pm$2.24 & \underline{0.94} & 5.44 & 60.27$\pm$4.51 & 0.77 & 4.88 & 66.89$\pm$9.48 & 0.18 & 5.69 & 54.51$\pm$12.79 & 0.18 & 2.80 & 38.88$\pm$2.57 & 0.28 \\
fedprox & & 6.46 & 54.25$\pm$12.36 & - & 6.84 & 64.60$\pm$8.42 & - & 7.02 & 59.98$\pm$7.74 & - & 6.71 & 45.30$\pm$10.26 & - & 7.18 & 58.00$\pm$8.46 & - & 5.29 & 55.21$\pm$12.92 & - & 4.73 & 36.19$\pm$3.14 & - \\
fedavg & & 6.80 & 53.84$\pm$14.42 & - & 7.96 & 62.22$\pm$10.06 & - & 7.93 & 56.11$\pm$9.67 & - & 7.73 & 43.82$\pm$11.51 & - & 5.07 & 67.05$\pm$6.94 & - & 5.78 & 53.56$\pm$13.41 & - & 6.57 & 31.69$\pm$5.57 & - \\
\bottomrule
\end{tabular}%
}
\end{table*}

\begin{table*}[ht]
\centering
\caption{Benchmarking results - No Quantity Skew scenario across datasets.}
\label{tab:no_skew_results}
\resizebox{\textwidth}{!}{%
\begin{tabular}{ll
cccc cccc cccc cccc cccc cccc cccc}
\toprule
Algorithm
& \multicolumn{3}{c}{All Datasets}
& \multicolumn{3}{c}{MNIST} 
& \multicolumn{3}{c}{Fashion-MNIST} 
& \multicolumn{3}{c}{KMNIST} 
& \multicolumn{3}{c}{CIFAR-10}
& \multicolumn{3}{c}{OctMNIST}
& \multicolumn{3}{c}{TissueMnist}\\
\cmidrule(lr){3-5} \cmidrule(lr){6-8} \cmidrule(lr){9-11} \cmidrule(lr){12-14} \cmidrule(lr){15-17} \cmidrule(lr){18-20} \cmidrule(lr){21-23}
& & Rank & Acc$\pm$Std & ARI 
& Rank & Acc$\pm$Std & ARI 
& Rank & Acc$\pm$Std & ARI 
& Rank & Acc$\pm$Std & ARI 
& Rank & Acc$\pm$Std & ARI 
& Rank & Acc$\pm$Std & ARI 
& Rank & Acc$\pm$Std & ARI \\
\midrule
cornflqs & & \textbf{2.45} & \textbf{73.62$\pm$14.38} & \textbf{0.96} & \underline{2.60} &  \textbf{89.65$\pm$2.85} & \textbf{1.00} & 3.20 & \textbf{82.69$\pm$1.25} & \textbf{1.00} & \underline{2.53} & \textbf{66.59$\pm$2.24} & \textbf{1.00} & \textbf{2.07} & \textbf{75.19$\pm$4.31} & \textbf{0.87} & \underline{2.67} & 66.42$\pm$5.00 & \textbf{0.87} & \textbf{1.50} & \textbf{41.84$\pm$0.68} & \textbf{1.00} \\
fl+hc & & \underline{2.94} & \underline{71.15$\pm$16.21} & \underline{0.84} & \textbf{2.53} & 89.48$\pm$3.08 & \textbf{1.00} & 3.20 & 82.55$\pm$1.37 & \textbf{1.00} & \textbf{2.00} & \underline{66.37$\pm$2.97} & 0.88 & \underline{2.47} & \underline{73.99$\pm$4.34} & \underline{0.73} & 3.13 & \textbf{66.77$\pm$3.14} & \underline{0.86} & 3.60 & 35.93$\pm$8.31 & \underline{0.74} \\
fedgroup & & 3.56 & 70.51$\pm$16.21 & 0.67 & 3.13 & \underline{89.64$\pm$2.72} & \textbf{1.00} & \textbf{1.47} & \textbf{82.84$\pm$1.29} & \textbf{1.00} & 4.07 & 66.15$\pm$2.72 & \underline{0.89} & 3.47 & 69.59$\pm$4.09 & 0.30 & 6.27 & 56.73$\pm$7.78 & 0.22 & \underline{3.20} & \underline{38.45$\pm$3.91} & 0.48 \\
ifca & & 3.64 & 68.02$\pm$16.55 & 0.80 & 3.25 & 89.58$\pm$2.68 & \underline{0.98} & \underline{2.73} & 82.04$\pm$2.34 & 0.94 & 3.07 & 66.53$\pm$2.25 & \textbf{1.00} & 6.53 & 60.45$\pm$2.35 & 0.71 &\textbf{ 2.00} & \textbf{67.10$\pm$3.22} & 0.75 & 4.00 & 36.57$\pm$4.51 & 0.40 \\
cfl & & 3.67 & 69.46$\pm$16.88 & 0.66 & 3.25 & 86.89$\pm$6.77 & 0.85 & 4.27 & 79.90$\pm$4.72 & 0.80 & 3.13 & 62.66$\pm$7.74 & 0.64 & 2.60 & 73.65$\pm$4.77 & 0.63 & 4.00 & 64.54$\pm$4.57 & 0.62 & 5.20 & 34.08$\pm$7.03 & 0.46 \\
srfca & & 5.82 & 64.23$\pm$17.12 & 0.44 & 6.47 & 82.69$\pm$11.23 & 0.69 & 5.47 & 80.52$\pm$1.58 & \underline{0.98} & 6.60 & 59.95$\pm$6.90 & 0.67 & 5.73 & 60.59$\pm$10.07 & 0.02 & 5.73 & 54.42$\pm$12.89 & 0.17 & 5.40 & 36.50$\pm$2.93 & 0.16 \\
fedprox & & 6.27 & 56.17$\pm$11.27 & - & 6.87 & 66.95$\pm$6.86 & - & 7.00 & 63.91$\pm$2.97 & - & 6.40 & 47.40$\pm$10.57 & - & 7.00 & 59.59$\pm$2.82 & - & 5.00 & 56.27$\pm$12.10 & - & 4.40 & 38.18$\pm$0.90 & - \\
fedavg & & 7.27 & 53.80$\pm$13.37 & - & 7.87 & 65.78$\pm$6.86 & - & 8.00 & 60.47$\pm$4.46 & - & 7.53 & 45.45$\pm$12.38 & - & 6.00 & 62.08$\pm$6.22 & - & 7.20 & 51.57$\pm$11.84 & - & 7.10 & 31.52$\pm$7.72 & - \\
\bottomrule
\end{tabular}%
}
\end{table*}

\begin{table*}[ht]
\centering
\caption{Benchmarking results - Quantity Skew Type 1 scenario across datasets.}
\label{tab:qs1_results}
\resizebox{\textwidth}{!}{%
\begin{tabular}{ll
cccc cccc cccc cccc cccc cccc cccc}
\toprule
Exp Type
& \multicolumn{3}{c}{All Datasets}
& \multicolumn{3}{c}{MNIST} 
& \multicolumn{3}{c}{Fashion-MNIST} 
& \multicolumn{3}{c}{KMNIST} 
& \multicolumn{3}{c}{CIFAR-10}
& \multicolumn{3}{c}{OctMNIST}
& \multicolumn{3}{c}{TissueMnist}\\
\cmidrule(lr){3-5} \cmidrule(lr){6-8} \cmidrule(lr){9-11} \cmidrule(lr){12-14} \cmidrule(lr){15-17} \cmidrule(lr){18-20} \cmidrule(lr){21-23}
& & Rank & Acc$\pm$Std & ARI 
& Rank & Acc$\pm$Std & ARI 
& Rank & Acc$\pm$Std & ARI 
& Rank & Acc$\pm$Std & ARI 
& Rank & Acc$\pm$Std & ARI 
& Rank & Acc$\pm$Std & ARI 
& Rank & Acc$\pm$Std & ARI \\
\midrule
cornflqs & & \textbf{1.87} & \textbf{74.88$\pm$15.23} & \textbf{0.91} & \textbf{1.88} & \underline{89.27$\pm$3.63} & 0.89 & \textbf{1.60} & \underline{83.44$\pm$1.12} & \textbf{1.00} & \textbf{2.13} & \textbf{65.70$\pm$2.52} & 0.89 & \textbf{1.47} & \textbf{80.99$\pm$4.00} & \textbf{0.84} & \underline{2.20} & \textbf{69.36$\pm$3.52} & \textbf{0.92} & \underline{2.30} & 39.44$\pm$1.25 & \textbf{0.91} \\
ifca & & \underline{2.91} & \underline{69.77$\pm$15.47} & \underline{0.78} & \underline{2.13} & \textbf{89.79$\pm$2.69} & \textbf{1.00} & \underline{1.80} & \textbf{83.47$\pm$1.01} & \textbf{1.00} & 2.60 & \underline{65.36$\pm$2.83} & \underline{0.92} & 6.47 & 65.91$\pm$2.74 & \underline{0.66} & \textbf{1.60} & \underline{69.02$\pm$3.84} & \underline{0.75} & 2.47 & \textbf{40.51$\pm$5.72} & 0.30 \\
fl+hc & & 4.15 & 67.26$\pm$15.80 & 0.44 & 4.07 & 78.16$\pm$11.70 & 0.45 & 4.93 & 75.56$\pm$6.30 & 0.57 & \underline{2.47} & 63.08$\pm$6.91 & 0.70 & 3.47 & 76.99$\pm$6.03 & 0.38 & 4.27 & 62.60$\pm$4.96 & 0.22 & 6.50 & 33.44$\pm$3.85 & 0.21 \\
cfl & & 4.57 & 66.51$\pm$15.40 & 0.38 & 5.73 & 76.85$\pm$6.21 & 0.37 & 4.87 & 75.84$\pm$6.86 & 0.56 & 5.40 & 55.97$\pm$6.10 & 0.36 & \underline{2.40} & \underline{78.83$\pm$5.77} & 0.41 & 4.13 & 64.90$\pm$4.39 & 0.27 & 5.80 & 34.25$\pm$3.59 & 0.22 \\
srfca & & 4.48 & 68.02$\pm$15.98 & 0.59 & 3.87 & 87.08$\pm$3.18 & \underline{0.90} & 4.60 & 78.76$\pm$1.98 & \underline{0.92} & 4.33 & 61.38$\pm$2.68 & \textbf{0.93} & 4.80 & 71.45$\pm$7.11 & 0.26 & 6.80 & 55.21$\pm$12.93 & 0.15 & \textbf{1.80} & \underline{39.80$\pm$1.61} & \underline{0.36} \\
fedgroup & & 4.85 & 66.17$\pm$16.90 & 0.38 & 3.73 & 86.70$\pm$5.53 & 0.73 & 3.13 & 81.61$\pm$3.21 & 0.81 & 4.40 & 61.35$\pm$6.40 & 0.66 & 5.53 & 67.43$\pm$8.33 & 0.02 & 6.27 & 56.20$\pm$10.75 & 0.03 & 6.00 & 34.29$\pm$2.71 & 0.12 \\
fedprox & & 6.53 & 56.11$\pm$12.35 & - & 6.93 & 65.63$\pm$7.45 & - & 7.00 & 62.67$\pm$4.41 & - & 6.67 & 46.11$\pm$11.38 & - & 7.33 & 62.01$\pm$5.91 & - & 5.80 & 56.19$\pm$13.45 & - & 3.90 & 36.91$\pm$1.51 & - \\
fedavg & & 6.53 & 56.01$\pm$15.36 & - & 8.12 & 61.81$\pm$9.99 & - & 8.00 & 56.81$\pm$9.26 & - & 7.87 & 44.37$\pm$12.92 & - & 4.53 & 72.22$\pm$4.54 & - & 4.93 & 56.11$\pm$14.89 & - & 6.50 & 31.69$\pm$6.18 & - \\
\bottomrule
\end{tabular}%
}
\end{table*}

\begin{table*}[ht]
\centering
\caption{Benchmarking results - Quantity Skew Type 2 scenario across datasets.}
\label{tab:qs2_results}
\resizebox{\textwidth}{!}{%
\begin{tabular}{ll
cccc cccc cccc cccc cccc cccc cccc}
\toprule
Algorithm
& \multicolumn{3}{c}{All Datasets}
& \multicolumn{3}{c}{MNIST} 
& \multicolumn{3}{c}{Fashion-MNIST} 
& \multicolumn{3}{c}{KMNIST} 
& \multicolumn{3}{c}{CIFAR-10}
& \multicolumn{3}{c}{OctMNIST}
& \multicolumn{3}{c}{TissueMnist}\\
\cmidrule(lr){3-5} \cmidrule(lr){6-8} \cmidrule(lr){9-11} \cmidrule(lr){12-14} \cmidrule(lr){15-17} \cmidrule(lr){18-20} \cmidrule(lr){21-23}
& & Rank & Acc$\pm$Std & ARI 
& Rank & Acc$\pm$Std & ARI 
& Rank & Acc$\pm$Std & ARI 
& Rank & Acc$\pm$Std & ARI 
& Rank & Acc$\pm$Std & ARI 
& Rank & Acc$\pm$Std & ARI 
& Rank & Acc$\pm$Std & ARI \\
\midrule
cornflqs & & \textbf{2.41} & \textbf{70.56$\pm$14.85} & \textbf{0.84} & \textbf{2.00} & \textbf{88.32$\pm$2.75} & \textbf{1.00} & \underline{2.13} & \textbf{82.18$\pm$1.32} & \textbf{1.00} & \textbf{1.40 }& \textbf{65.29$\pm$2.15} & \textbf{1.00} & \textbf{2.13} & \underline{72.74$\pm$4.40} & \textbf{0.64} & 4.33 & 61.78$\pm$2.98 & \underline{0.67} & 3.20 & \underline{38.98$\pm$0.48} & \textbf{0.66} \\
fl+hc & & \underline{2.59} & \underline{70.02$\pm$15.44} & \underline{0.77} & \underline{2.33} & \underline{88.04$\pm$3.00} & \textbf{1.00} & 2.67 & \underline{81.98$\pm$1.65} & \textbf{1.00} & \underline{2.33} & \underline{65.05$\pm$2.36} & \textbf{1.00} & \underline{2.27} & \textbf{76.83$\pm$6.44} & 0.53 & \textbf{2.60} & \textbf{64.71$\pm$3.74} & \textbf{0.83} & \underline{3.10} & 37.65$\pm$3.65 & 0.38 \\
fedgroup & & 3.93 & 66.12$\pm$16.97 & 0.62 & 2.87 & 87.83$\pm$3.50 & \underline{0.97} & \textbf{1.87} & 81.95$\pm$1.79 & \underline{0.98} & 3.07 & 63.86$\pm$3.28 & 0.83 & 4.33 & 67.88$\pm$5.25 & 0.11 & 6.27 & 55.81$\pm$6.88 & 0.44 & 4.40 & 36.08$\pm$2.61 & \underline{0.60} \\
ifca & & 3.87 & 64.92$\pm$17.67 & 0.66 & 3.00 & 86.97$\pm$5.01 & 0.93 & 3.67 & 80.13$\pm$2.73 & 0.86 & 3.47 & 63.60$\pm$3.55 & \underline{0.84} & 6.47 & 61.20$\pm$1.18 & \underline{0.61} & \underline{3.20} & \underline{63.55$\pm$1.48} & 0.58 & 4.00 & 31.57$\pm$7.01 & 0.07 \\
srfca & & 4.29 & 65.14$\pm$15.49 & 0.55 & 4.80 & 84.45$\pm$3.01 & 0.78 & 4.93 & 77.48$\pm$2.07 & 0.90 & 5.40 & 59.48$\pm$2.66 & 0.71 & 3.82 & 69.27$\pm$7.06 & 0.26 & 4.53 & 53.91$\pm$13.41 & 0.22 & \textbf{1.20} & \textbf{40.34$\pm$0.72} & 0.32 \\
cfl & & 5.36 & 60.40$\pm$13.48 & 0.34 & 6.33 & 71.43$\pm$7.44 & 0.40 & 5.80 & 68.10$\pm$9.34 & 0.60 & 5.47 & 54.42$\pm$6.41 & 0.48 & 3.80 & 69.24$\pm$4.65 & 0.28 & 4.73 & 59.70$\pm$5.78 & 0.25 & 6.60 & 31.75$\pm$1.78 & 0.03 \\
fedprox & & 6.59 & 50.25$\pm$12.64 & - & 6.73 & 61.22$\pm$10.06 & - & 7.07 & 53.36$\pm$9.43 & - & 7.07 & 42.39$\pm$8.68 & - & 7.20 & 52.41$\pm$11.35 & - & 5.07 & 53.18$\pm$13.80 & - & 5.90 & 33.48$\pm$3.94 & - \\
fedavg & & 6.61 & 51.55$\pm$14.23 & - & 7.93 & 59.09$\pm$12.12 & - & 7.80 & 51.05$\pm$11.83 & - & 7.80 & 41.66$\pm$9.37 & - & 4.71 & 66.48$\pm$6.08 & - & 5.20 & 53.01$\pm$13.84 & - & 6.10 & 31.86$\pm$1.46 & - \\
\bottomrule
\end{tabular}%
}
\end{table*}

Once stability is reached, Algorithm~\ref{alg:FedAvg} takes over using the final clustering $K^{(r)}$ obtained from Algorithm~\ref{alg:client-side}. The server continues to aggregate models within each cluster, broadcast them, and collect client updates. Since cluster memberships $k_i^*$ remain fixed, the server simply sends each client its corresponding model $W^{(k_i^*)}$ (line 6). Clients then perform local training and return their updates without further loss evaluations. The algorithm terminates when the predefined number of rounds $N$ is reached. It is worth noting that if Algorithm~\ref{alg:client-side} never reaches stability, Algorithm~\ref{alg:FedAvg} will not proceed, as the maximum number of rounds $N$ would have already been completed at the end of Algorithm~\ref{alg:client-side}. This complete workflow is designed to ensure that CORNFLQS automatically adapts to both QS and non-QS scenarios without the need for explicit QS detection, while aiming to maintain competitive performance against state-of-the-art CFL methods in Non-IID settings.

\section{Experimental results and discussion on CORNFLQS}\label{sec:cornflqs-eval}

This section compares CFL algorithms to evaluate the \textbf{effectiveness}, \textbf{robustness}, and \textbf{sensitivity} of CORNFLQS. All experiments uses the same setup as Section~\ref{sec:CFLQS}. We add FedAvg \cite{FedAvg} and FedProx \cite{fedprox} as baselines too. CORNFLQS is compatible with any clustering method and similarity metric. In our experiments, we used agglomerative hierarchical clustering with Euclidean distance and Ward linkage, chosen for its ability to minimize intra-cluster variance. Our analysis is structured around three research questions:
\begin{itemize}[label=\textbullet, left=0pt, itemsep=0.5ex]
    \item \textbf{Effectiveness} — \textit{How does CORNFLQS perform compared to other CFL algorithms?}
    Effectiveness is assessed through average accuracy, ARI, and standard deviations across all setups.
    
    \item \textbf{Robustness} — \textit{How consistent is CORNFLQS across Non-QS, QS1, QS2, and non-IID settings?}
    It is measured using the average rank of each model across experiments.
    
    \item \textbf{Sensitivity} —  \textit{How does performance change with the number of clusters?}
     We test varying cluster numbers for CORNFLQS.
\end{itemize}

\subsection{Effectiveness}

The results in Tables~\ref{tab:global_results} to \ref{tab:qs2_results} highlight the \textbf{performance} but also \textbf{robustness} of CORNFLQS across benchmarks and datasets. Table~\ref{tab:global_results} summarizes average results from Tables~\ref{tab:no_skew_results} to \ref{tab:qs2_results}, reporting accuracy, ARI, both with standard deviations and average ranking for each dataset. Rankings are computed by ordering algorithms by accuracy in each setup, then averaging these ranks for a given dataset. While no method dominates every setup, CORNFLQS regularly ranks first or second for both accuracy and ARI, confirming its strong and consistent performance over standard and CFL baselines. Notably, it achieves the best global average rank (2.24), outperforming methods like FedAvg and FedProx. It is interesting to note that model architecture choice \cite{medmnist} are underfitting on TissueMnist but CORNFLQS still outperforms other methods globally on this dataset.

\begin{figure}[ht]
\includegraphics[width=0.75\linewidth]{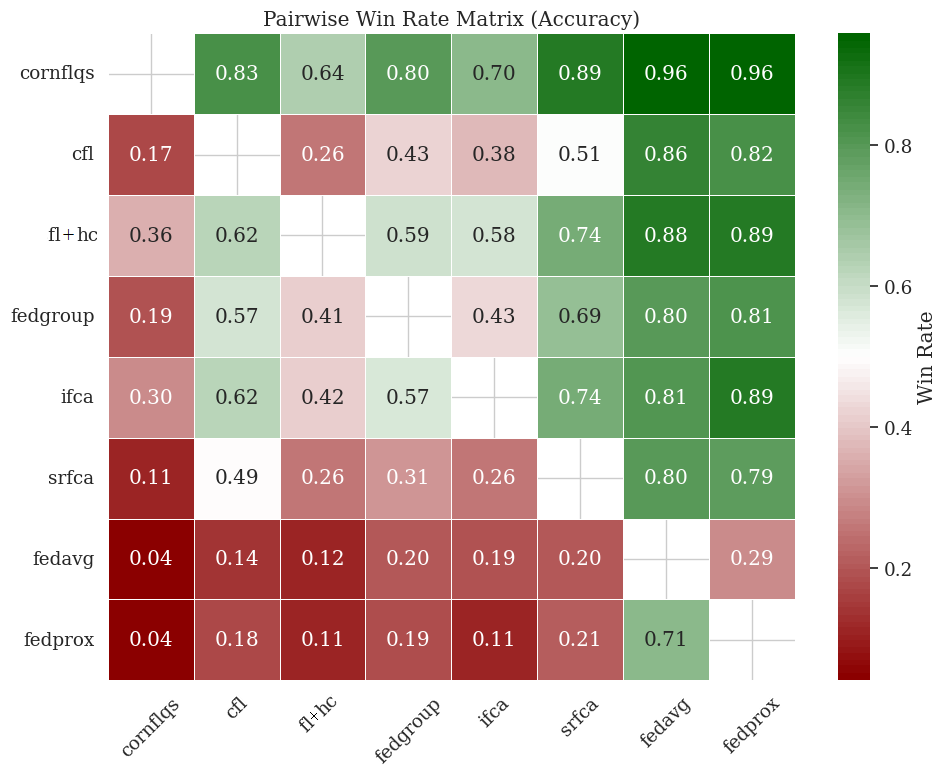}
\Description{}
\caption{Winrate matrix based on accuracy across all experiments of CFL algorithms}
\label{fig:winrate}
\end{figure}

Lastly, Figure~\ref{fig:winrate} visualizes pairwise win rates between algorithms. Each row and column represents a CFL algorithm, and the win rate corresponds to the percentage of times the algorithm in the row outperforms the one in the column across all experiments. To improve readability, win rates above 50\% are shaded green (positive win rates) and those below 50\% in red (negative win rates). As expected, CORNFLQS consistently maintains a positive win rate against all competitors, with its closest challengers being FL+HC and IFCA. Interestingly, no algorithm has a 0\% win rate — not even FedAvg — indicating that even basic methods can occasionally outperform CFL algorithms in particular cases if clustering fails.

\subsection{Robustness}

It is important to note that an algorithm can have a better average ranking but a lower average accuracy, and vice versa. This highlights a crucial nuance: a method might deliver high accuracy in specific cases but lack robustness and be consistently outperformed in others. The average ranking metric, especially when computed across diverse experimental conditions, captures broader algorithmic resilience and consistency.

In practice, CORNFLQS demonstrates strong robustness across all QS scenarios. Globally, it always achieves the best average rank: 2.24 in the global setting, 2.45 under no QS, 1.87 in QS1, and 2.41 in QS2 (Table \ref{tab:global_results}). These results reflect its stability regardless of datasets or experimental configurations. While occasional fluctuations occur—where other methods may outperform it on specific datasets and/or QS scenarios—CORNFLQS consistently secures a top position overall. With the exception of OctMNIST in QS1 and FashionMNIST in the non-QS setting, where CORNFLQS ranks third, it is consistently first or second in average ranking and dominates in most datasets and QS cases. This is further highlighted by the average $\Delta$ ARI results. CORNFLQS achieves a $\Delta$ ARI of $0.05$ for QS1 and $0.13$ for QS2 on average across all datasets. When compared to the values observed in the "All Datasets" columns of the ARI discrepancy heatmaps (Figure~\ref{fig:delta_metrics}), these results indicate that CORNFLQS achieves competitive discrepancy between non-QS and QS scenarios against all considered algorithms.

To further assess robustness across different data heterogeneity scenarios of CORNFLQS, Table~\ref{tab:cfl_heterogeneity_results} reports average accuracy rankings for different heterogeneity types under various QS settings. CORNFLQS once again displays robust behavior across most heterogeneity types. In particular, with (a) concept shift on features (CSF) and (c) feature distribution skew (FDS), it achieves the best average ranking across all experiments. It ranks third only in (b) concept shift on labels (CSL) in the non-QS setting, and second in QS2, while maintaining first place in QS1.

\begin{table}[ht]
\small
\centering
\caption{Average accuracy ranking by heterogeneity types with (a) Concept Shift on Features (CSF), (b) Concept Shift on Labels (CSL), and (c) Feature Distribution Skew (FDS).}
\label{tab:cfl_heterogeneity_results}
\resizebox{\linewidth}{!}{%
\begin{tabular}{lccc|ccc|ccc}
\toprule
\multirow{2}{*}{Algorithm} & \multicolumn{3}{c|}{NoQS} & \multicolumn{3}{c|}{QS1} & \multicolumn{3}{c}{QS2} \\
& (a) CSF & (b) CSL & (c) FDS & (a) CSF & (b) CSL & (c) FDS & (a) CSF & (b) CSL & (c) FDS \\
\midrule
cornflqs & \textbf{2.43} & 2.63 & \textbf{2.30} & \textbf{1.66} & \textbf{1.93} & \textbf{2.07} & \textbf{2.23} & \underline{2.68} & \textbf{2.40} \\
FL+HC     & 3.80 & \textbf{2.03} & \underline{2.83} & 5.20 & \underline{2.17} & 4.90 & 3.83 & \textbf{1.00} & \underline{2.47} \\
ifca     & 3.74 & 3.58 & 3.60 & \underline{3.49} & 2.91 & \underline{2.23} & 4.40 & 3.17 & 3.97 \\
fedgroup & \underline{3.60} & 3.90 & 3.17 & 4.54 & 5.17 & 4.90 & 4.20 & 4.08 & 3.50 \\
srfca    & 4.94 & 6.47 & 6.20 & 3.80 & 5.53 & 4.23 & \underline{3.67} & 5.20 & 4.21 \\
cfl      & 3.94 & \underline{2.23} & 4.77 & 4.49 & 4.37 & 4.87 & 5.31 & 5.12 & 5.60 \\
fedprox  & 5.97 & 6.67 & 6.23 & 6.31 & 6.67 & 6.63 & 6.14 & 7.00 & 6.77 \\
FedAvg   & 7.17 & 7.76 & 6.90 & 6.50 & 6.93 & 6.17 & 5.97 & 7.32 & 6.76 \\
\bottomrule
\end{tabular}%
}
\end{table}

\subsection{Sensitivity to clustering hyperparameter}
\begin{figure}[ht]
\includegraphics[width=0.6\linewidth]{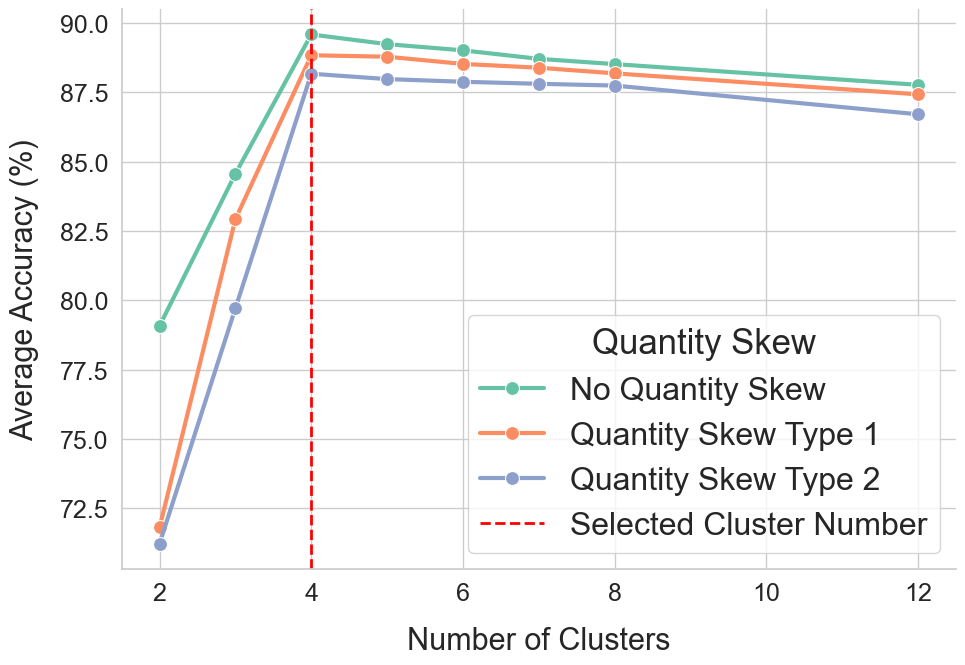}
\Description{}
\caption{Impact of Cluster Number on Average Accuracy of CORNFLQS for MNIST}
\label{fig:num_clusters}
\end{figure}

A key hyperparameter of CORNFLQS is the number of clusters ($K$). In our experiments, we use $K = 4$ clusters, reflecting the $4$ heterogeneity classes from the benchmark. In practice, this value is hard to determine without data or expert domain knowledge and is typically chosen by grid search. Most CFL algorithms holds this hyperparameter too (e.g., CFL, IFCA, FedGroup), while others, like SRFCA and FL+HC, rely on threshold-based metrics (FL+HC, for example, can use either \cite{FL+HC}). 
To test CORNFLQS’s sensitivity to this parameter, we tested varying the number of clusters with MNIST dataset across different heterogeneity types and QS scenarios (Similar behavior was observed across other datasets evaluated, but these results are omitted from the article due to space limitation). Figure~\ref{fig:num_clusters} reports the average accuracy for Non-QS (green), QS1 (orange), and QS2 (blue) for various values of number of clusters. As expected, the best accuracy occurs at $4$ clusters. It is interesting, that overestimating the number of clusters degrades performance much less than underestimating it. This shows that CORNFLQS is robust to overestimation but sensitive to underestimation (will group dissimilar client, harming performances). This underlines the importance of careful cluster count selection or favoring overestimation.

\section{Conclusion}\label{sec:conclusion}
 In this paper, we investigate Quantity Skew (QS) in Clustered Federated Learning (CFL), showing with 270 experiments that QS in multiple non-IID contexts poses a significant challenge to existing FL and CFL methods. To address this issue, we developed CORNFLQS, a new CFL algorithm that combines weight- and loss-based clustering. This approach delivers both effectiveness and robustness to QS scenarios, surpassing the performance of existing state-of-the-art algorithms. Future work should include extending CORNFLQS to larger and more complex models and Non-IID settings, while leveraging its efficacy in combination with FL techniques such as differential privacy and client selection. Evaluating CORNFLQS with alternative clustering algorithms and metrics may further improve its efficiency in specific scenarios. Additional studies could also investigate computational and communication costs in practical deployments and refine methods for selecting the number of clusters to enhance real-world applicability.

\newpage
\section*{GenAI Usage Disclosure}

During the preparation of this work, the authors used generative AI tools in a limited and supportive capacity:
\begin{itemize}
    \item \textbf{ChatGPT 4.0 (OpenAI)} was used to assist in improving the formulation, phrasing, and clarity of the English text, as well as edit and polish authors’ work. No new scientific content, novel results, or sections of the paper were generated by the tool. All final text was written, verified, and approved by the authors.
    \item \textbf{GitHub Copilot (OpenAI)} was used to assist with code refactoring, provide inline code completion suggestions, and generate documentation comments during software development. All code incorporated in the final version was reviewed and validated by the authors to ensure accuracy and alignment with the research objectives.
\end{itemize}

The authors confirm that all scientific claims, results, experiments, and conclusions presented in this paper are their own.

\bibliographystyle{ACM-Reference-Format}
\balance
\bibliography{biblio}

\end{document}